# Generation of Standardized E-Learning Contents from Digital Medical Collections

Felix Buendía[1], Joaquín Gayoso-Cabada[2], José-Luis Sierra[2]

[1]*Escuela Técnica Superior de Ingeniería Informática, Universitat Politècnica de València. Valencia, Spain*
fbuendia@disca.upv.es

[2]*Facultad de Informática. Universidad Complutense de Madrid. Madrid. Spain*
{jgayoso, jlsierra }@ucm.es

*Abstract*

In this paper, we describe an approach to transforming the huge amount of medical knowledge available in existing online medical collections into standardized learning packages ready to be integrated into the most popular e-learning platforms. The core of our approach is a tool called *Clavy*, which makes it possible to retrieve pieces of content in medical collections, to transform this content into meaningful learning units, and to export it in the form of standardized learning packages. In addition to describing the approach, we demonstrate its feasibility by applying it to the generation of IMS content packages from *MedPix*, a popular online database of medical cases in the domain of radiology.

## Introduction

In recent years, there has been a proliferation of online digital medical collections that contribute to organizing medical knowledge in many different areas either for clinical research or education purposes [1–4]. All these collections have undoubted educational value, as evidenced by multiple experiences on the use of data repositories for medical education [5–8]. However, many times these knowledge sources require suitable organization to support instruction. In addition, it would be convenient to have systematic methods that allow the integration of these practices in standard e-learning scenarios. In this paper, we address these concerns by presenting one of these methods, which makes it possible to produce standardized e-learning content packages from these kinds of digital collections.

Our method is supported by a tool called *Clavy* [9–11], which facilitates the importation of data and content from existing medical collections, the transformation and reorganization of all this information to meet specific training criteria, and the exportation of the resulting structures as standardized content packages. In order to demonstrate the feasibility of the approach, we will describe its application to *MedPix* [12], a medical digital library integrating a huge amount of medical cases in the field of radiology. In this way, we will show how, by using *Clavy*, it is possible to generate IMS CP content packages [13], which can be integrated in widely-used e-learning platforms (Moodle, in particular). Preliminary stages of this work are described in [14, 15].

## From Existing Medical Collections to Standardized Learning Content

In this section, we describe the generic process model followed by our approach, and how this process model is supported by the *Clavy* tool.

**The Generation Process Model**

The activity diagram in Figure 1 describes the process model adopted by our approach. This model orchestrates the collaboration of two different actors (represented as separate swim lanes in the activity diagram) for the production of standardized medical learning content:

- *Developers*, who are in charge of providing the tools necessary for transforming the contents offered by existing medical collections into standard learning packages during *development* activities: *importation / reconfiguration / exportation engine development* activities.

- *Domain experts*, who, in turn, use the tools provided by developers to actually generate the learning packages during *generation* activities: *medical collection importation / reconfiguration* and *learning content exportation* activities.

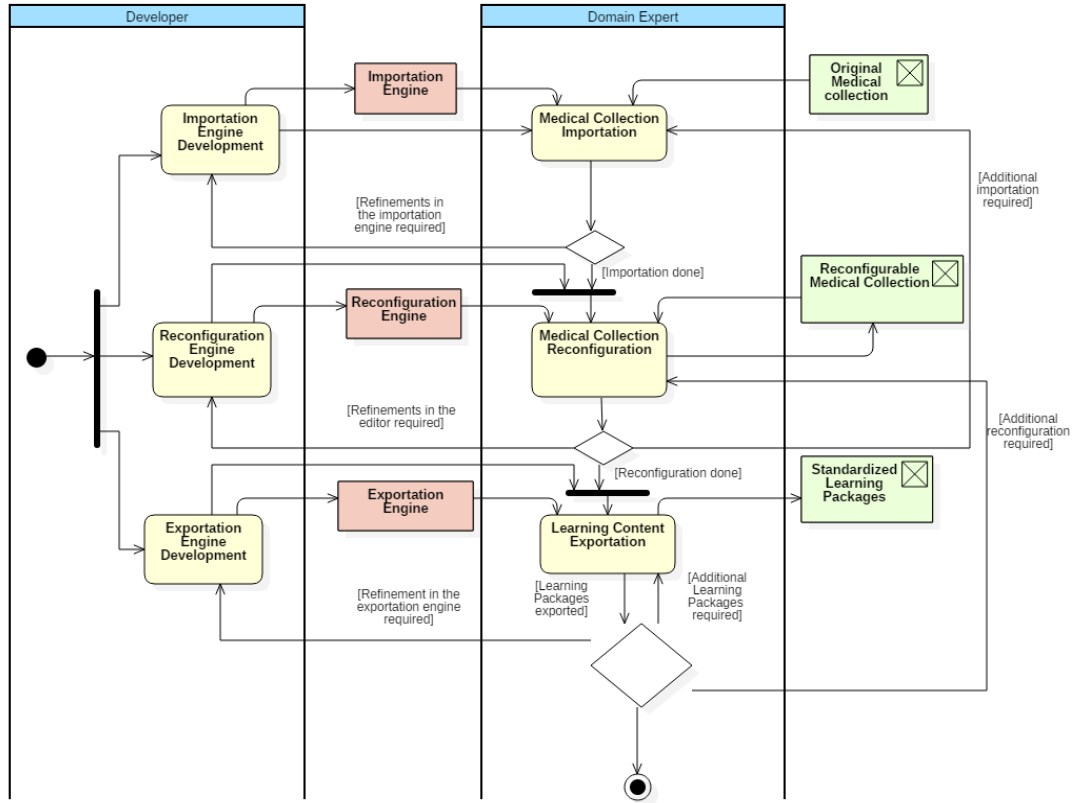

Figure 1. The generation process model

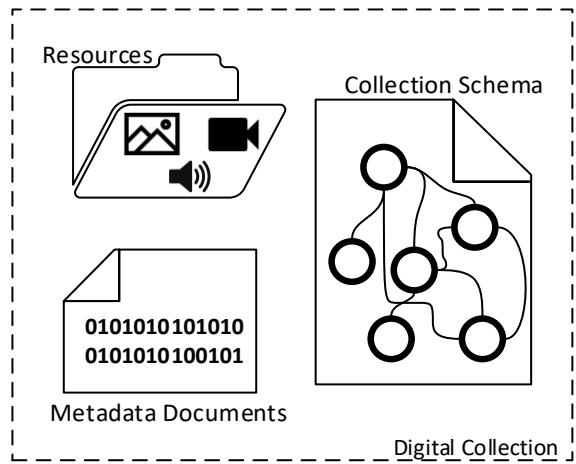

Figure 2. Canonical representation of a medical collection

In addition, a keystone concept in the process model is the *canonical representation* of medical collections. Such canonical representation conceives of a medical collection as one made of (Figure 2):

- A set of *resources* (multimedia files, video, audio, text, etc.), which represent the local contents of the collection, as well as external contents represented by their URLs.
- A set of *metadata documents* that describe how to put chunks of resources together in order to yield significant knowledge units, and which also provide additional meta-information concerning these knowledge units.
- A *metadata scheme*, which is an explicit representation of the structure followed by the metadata documents.

In this way, the steps of the generation workflow consist of:

- The input to the generation model is an existing medical collection, identified as an *original medical collection* in Figure 1, representing the primary source of medical knowledge and content that will be exploited to produce the learning packages.
- By using a suitable *importation engine*, which is provided by developers, domain experts are able to mirror the original medical collection in terms of the canonical representation to yield a *reconfigurable medical collection*. To do so, the importation engine can provide several customization mechanisms, allowing experts to decide the specific information to be imported and to fine-tune the way in which this information will be reflected in the reconfigurable collection.
- Then, domain experts can transform this medical collection with a suitable *reconfiguration engine*, also provided by developers, to make its pedagogical value explicit. The reconfiguration engine will let domain experts edit not only the resources and the metadata documents, but also the metadata scheme, in order to better adapt the organization of the collection to specific educational settings.
- Finally, developers can offer domain experts a suitable *exportation engine*, which lets them generate standardized learning packages from the reconfigurable medical collection. As with the importation engine, the exportation engine can also provide a set of features that the domain expert can customize to generate and package the learning content (e.g., when a content package contains a single learning unit or when it contains several, how this content is organized, etc.).

Finally, it is worthwhile to notice that the process model is not a sequential one but is iterative and incremental in nature. Indeed, once domain experts use a tool provided by developers, they may discover some shortcomings in this tool (e.g., some limitations, missed requirements, the need for better ways of performing a customization or editing task, etc.). In consequence, instead of continuing with the usual generation workflow, domain experts can return the tools to developers in order to let them resolve the shortcomings detected. In addition, domain experts themselves can backtrack to previous generation activities to solve any omission or any aspect not properly addressed, which introduces new loops in the generation process.

**Supporting the Generation Process Model with *Clavy***

The aforementioned *Clavy* tool has been designed to support the process model described earlier. For this purpose:

- *Clavy* supports a canonical representation for medical collections like that described in the previous subsection and displayed in Figure 2. Metadata documents in *Clavy* represent logical aggregations and organizations of resources in terms of *element-value* pairs. *Clavy* envisions four different kinds of elements: (i) *descriptive* elements, whose values can be text-based descriptions associated to the resources; (ii) *resource* elements, whose values are references to the resources in the collection; (iii) *link* elements, whose values are references to other documents (thus making it possible to build collection-specific semantic webs); and (iv) *structural* elements, whose only purpose is to enhance document organization and readability. In addition, metadata schemata in *Clavy* describe the hierarchical organization of elements in the collections' documents.
- *Clavy* is equipped with an extensible importation system supported by an *importation plug-in architecture*. Therefore, developers can provide importation engines as *importation plug-ins* that can be plugged in to *Clavy* for their use by domain experts. These importation plug-ins can be general-purpose ones, enabling the importation of data from standard platforms and formats (XML, JSON, relational databases, etc.), or they can be targeted to specific collections (e.g., a *plug-in* to import data from a medical collection using a specific RESTfull API). This feature lets domain experts retrieve pieces of content that are usually poorly structured in external sources (e.g., external medical collections) and arrange them in the versatile and powerful organization provided by the canonical representation. Figure 3a shows a screenshot of an importation plug-in for getting data from *MedPix*, a medical collection containing medical cases in the Radiology domain (see next section).

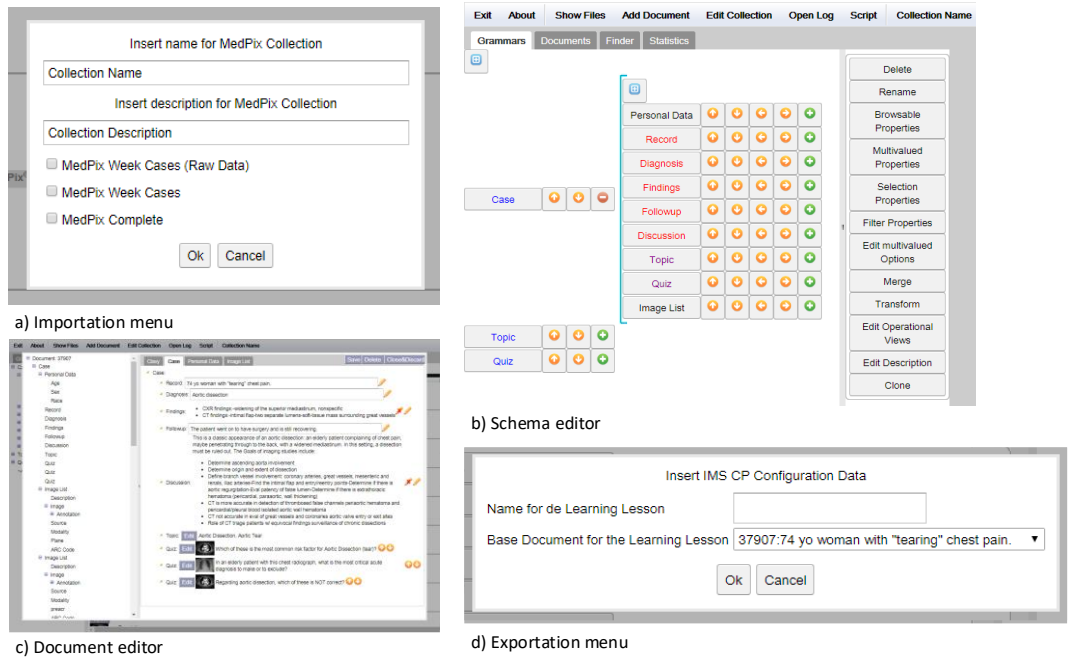

Figure 3. Snapshots of the *Clavy* platform

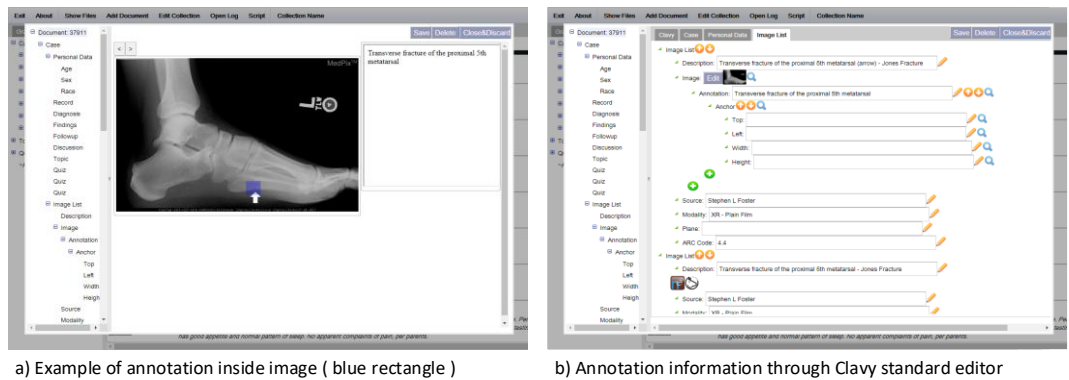

Figure 4. An edition plug-in in action

- *Clavy* provides an extensible reconfiguration engine. On the one hand, this engine integrates a user-friendly editor for metadata schemata oriented to domain experts, which does not assume any specific knowledge on data management or computer science (see Figure 3b). Indeed, since all the documents in a *Clavy* collection share the common structure described by its schema, *Clavy* lets domain experts reconfigure this structure by editing only the schema. For this purpose, the editor for the metadata schemata lets them: (i) rename element types; (ii) remove useless or non-relevant elements; (iii) merge two semantically-equivalent element types; and (iv) change the hierarchical organization of elements. Once the schema is reconfigured, *Clavy* takes care of the required adjustments in the internal collection's organization in a seamless and automatic way. In addition to the schema editor, the reconfiguration engine integrates an extensible document editor, which lets domain experts tweak individual documents in order to carry out more in-depth content-specific curation activities. By default, this editor is a form-oriented one (see Figure 3c). However, *Clavy* makes it possible to customize the edition flow by introducing another plug-in architecture: the *edition plug-in architecture*. This architecture lets developers specialize the reconfiguration engine to the specific needs of domain experts by providing plug-ins able to take control of the edition of particular parts of the document. For instance, Figure 4a exemplifies one of these specialized plug-ins in action to annotate medical images with explanations about relevant findings in the image, and constitutes an example of the way in which *Clavy* lets domain experts enrich content imported from external medical collections. Figure 4b also shows the same information edited using the default reconfiguration engine.
- Finally, to support exportation, *Clavy* provides a third plug-in architecture: the *exportation plug-in architecture*. Using this architecture, developers can plug exportation engines in to *Clavy* as exportation plug-ins, and domain experts can access these plug-ins to generate the learning packages. For instance, Figure 3d shows a user interface of an exportation plug-in for generating leaning content packaged as IMS CP packages.

## From *MedPix* to IMS CP Content Packages

In this section, we demonstrate how *Clavy* is used to import information from *MedPix*, to reconfigure the resulting canonical collection, and to export content that conforms to the IMS CP specification.

### Importation

As indicated in the previous section, the first step to take in order to let *Clavy* import content from an external medical collection is to equip it with a suitable importation *plug-in*. In the case of *MedPix*, it is a collection-specific *plug-in*, which is provided by the developers during the *importation engine development* activity. Figure 5 sketches the internal organization of this importation *plug-in*, which is able to ingest the relevant *MedPix* clinical cases, along with all the associated information. In *MedPix*, *clinical cases* include clinical images and additional descriptive information and cover different *clinical topics*, since the two types of elements are cross-referenced. Therefore, once the clinical cases to be processed are indicated by the domain expert (during the *medical collection importation* activity; see the snapshot of the plug-in's user interface in Figure 3), the following steps are performed:

- The *plug-in* uses the *MedPix* REST API to recover the URLs in *MedPix* for these clinical cases.
- In turn, each case can be recovered by using the REST API again. Then, by scraping each case, the *plug-in* is able to discover the set of related topics. The actual information for the topics can be retrieved by using the REST API a third time.
- Topics are in turn scraped to retrieve additional related cases, which are then ingested and analyzed until all the relevant information has been retrieved.
- Once all the relevant information is ingested, the *plug-in* makes all this information persistent as a *Clavy* canonical collection.

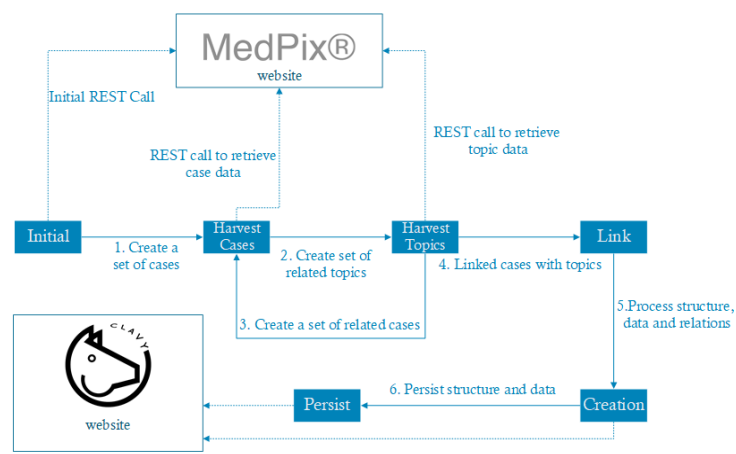

Figure 5. Internal organization of the *Clavy* importation plug-in for *MedPix*

Figure 6 shows an excerpt of the schema for this collection. This schema contains 72 elements, and it mirrors the basic *MedPix* structure in *Clavy* terms:
- The *Case* element groups together all the information that concerns clinical cases. Among other aspects, this element contains data related to the patient associated with the clinical case (*sex*, *age*, *diagnosis*, etc.), the *encounters* held between physicians and patient, links to topics related to the case, and clinical images associated with such a case (for each image, besides the caption and the URL, technical data similar to that supported by formats such as DICOM is also provided).

- The *Topic* element contains all the information concerning a medical topic. Among other aspects, it supports links to related cases, description of related topics, classification in terms of keywords, the ACR Code [16], categories and subcategories, etc.

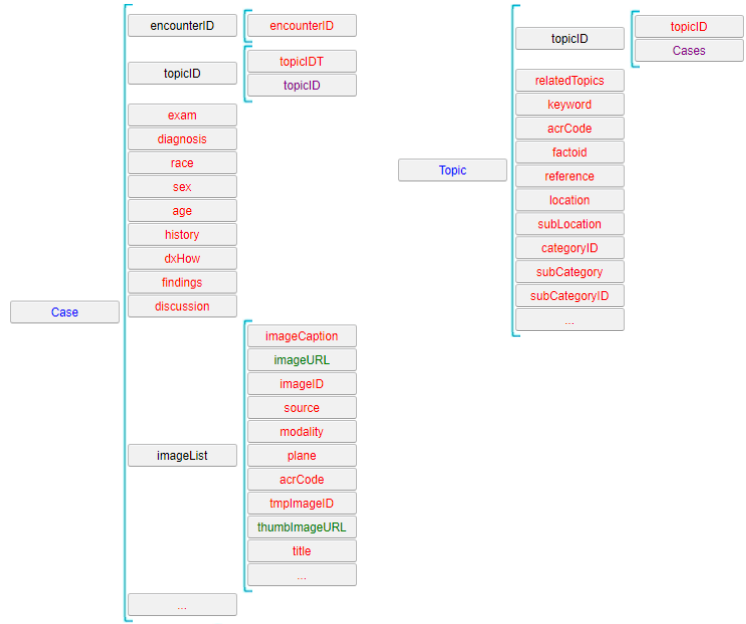

Figure 6. Excerpt of the initial *Clavy* schema for the contents imported from *MedPix*

**Reconfiguration**

Once the canonical collection for *MedPix* is made available in *Clavy*, domain experts can reconfigure this collection by using the schema and document editors (*medical collection reconfiguration* activity).

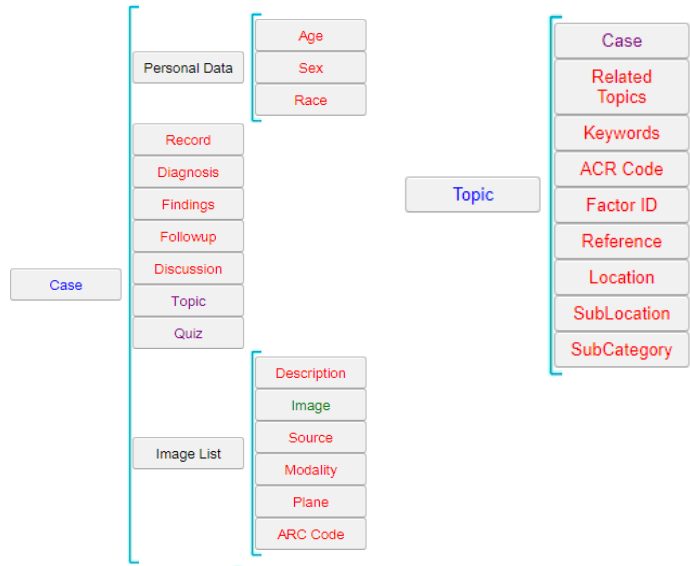

Figure 7. Reconfigured *Clavy* schema for the contents imported from *MedPix*

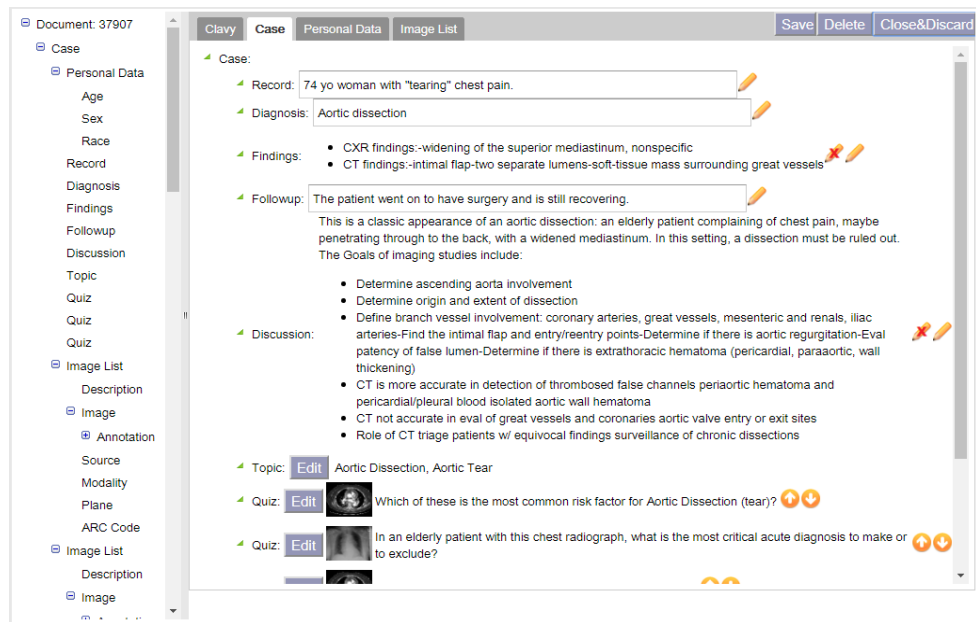

Figure 8. A document for a *MedPix* clinical case edited in *Clavy*

Most of the reconfiguration process can be achieved by reorganizing the collection schema. Indeed, as indicated earlier, the original schema involved 72 elements, many of which are not excessively interesting from an educational point of view. In addition, the organization of the schema was susceptible to improvement to better clarify the different aspects addressed by the cases and topics. Hence, the schema was tweaked by domain experts to:

- Suppress unnecessary information; for instance, elements like internal *MedPix* IDs used for cross-reference and internal organization, or detailed properties of the images that hardly had any educational value, and therefore were erased from the schema.
- Promote a better structure for the documents. For instance, all the information concerning the patient in a clinical case was grouped behind a *Personal Data* structural element, elements were renamed with more descriptive names, etc.

Figure 7 shows the resulting *Clavy* schema after these reconfigurations. The 72 initial elements were reduced to 28, the most useful from an educational point of view, plus some oriented to enhancing structure (e.g., the aforementioned *Personal Data* structural element). Finally, it is worthwhile to point out that, as indicated earlier, all the changes required by these reconfigurations in the underlying repository were automatically carried out by the tool.

In addition to editing the schema, instructors were also able to edit individual documents to inspect how changes in the schema propagate and to curate contents. Figure 8 shows a screenshot of the document editor acting on a document sample associated with a *MedPix* clinical case. As well, the document editor was previously customized by developers with an edition *plug-in* for annotating images (*reconfiguration engine development* activity; Figure 4a actually shows an screenshot of this plug-in's user interface). It allowed domain experts to enrich the *MedPix* contents with multimedia annotations of the images integrated in the clinical cases.

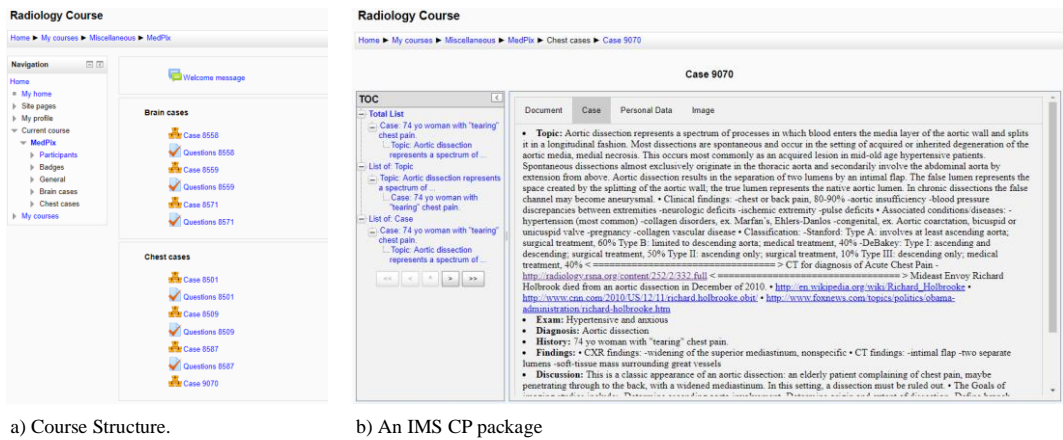

Figure 9. A *MedPix* clinical case exported by *Clavy* as an IMS CP package

| a) Course Structure. | b) An IMS CP package |

Figure 10: A Moodle course that integrates *MedPix* clinical cases, reconfigured in *Clavy*, and packaged as IMS CP content packages

**Exportation**

By providing an appropriate exportation plug-in during the *exportation engine development* activity, developers let domain experts export *Clavy* content as standardized learning packages. This is the case of the IMS CP exportation *plug-in* currently available in *Clavy*. By using this *plug-in*, domain experts can easily address the *learning content exportation* activity to produce IMS CP packages from *Clavy* collections. This *plug-in* accepts a set of documents to be exported as input (a screenshot of its user interface is shown in Figure 3), and recursively follows the structure imposed by the link elements to recreate these contents in IMS CP packages. In order to do so, it uses additional attributes attachable with elements to decide where to represent descriptive elements as organization items or as additional IMS CP resources. It also monitors and breaks potential cycles in the link structure, since the IMS CP information model is primarily tree-shaped.

Figure 9 illustrates the effect of this exportation *plug-in* on a *MedPix* clinical case reconfigured in *Clavy*, highlighting the structure of the resulting package, as well as an excerpt of the IMS CP manifest. The resulting packages can be readily incorporated into any learning management system that supports the IMS CP standard. For instance, Figure 10a shows a screenshot of a sample radiology course deployed in Moodle, which interleaves clinical cases incorporated as IMS CP content packages generated from *Clavy* with quizzes on the cases also taken from *MedPix*. Figure 10b illustrates one of these packages displayed in a Moodle platform.

# Conclusions and Future Work

In this work, we have described an approach to the production of standardized learning content from existing medical collections. The approach presented promotes the importation of data and content from the collections and the recreation of all this information in a canonical collection. Then, this collection can be adequately reconfigured and adapted to the specific formative needs of particular training scenarios. Finally, the assets in the reconfigured collection can be exported and packaged using widely used e-learning standards. We have shown how this approach is implemented using an information management tool called *Clavy*, and we have demonstrated its feasibility with a case study concerning the importation, reconfiguration and exportation as IMS CP content packages of clinical cases coming from *MedPix*, a popular medical collection in the field of radiology.

Currently we are working on supporting the exportation of *Clavy* documents to other e-learning formats that support interaction (e.g., in particular, SCORM packages) [17]. We are also working on the importation of MCQs from *MedPix* and the embedment of these MCQs in SCORM packages. Further works plan to support the exploitation of *Clavy* aggregation features to define learning paths, the exportation of content in terms of the IMS *Common Cartridge* recommendation [17], and the assessment of courses integrating learning content generated through our approach in residency hospital programs.

## Compliance with Ethical Standards

Funding: This study was funded by the Ministry of Science, Innovation and Universities, Spain (grants numbers TIN2014-52010-R and TIN2017-88092-R).

Conflict of Interest: Félix Buendía declares that he has no conflict of interest. Joaquín Gayoso-Cabada declares that he has no conflict of interest. José-Luis Sierra declares that he has no conflict of interest.

Ethical approval: This article does not contain any studies with human participants or animals performed by any of the authors.